*Research Article*

# Methodological Precedence in Health Tech: Why ML/Big Data Analysis Must Follow Basic Epidemiological Consistency. A Case Study

Marco Roccetti[1,*]

[1] Department of Computer Science and Engineering, University of Bologna, Bologna, Italy

**\*  Correspondence:** Email: marco.roccetti@unibo.it; Tel: +393920271318.

**Abstract:** The integration of advanced analytical tools, including Machine Learning (ML) and massive data processing, has revolutionized health research, promising unprecedented accuracy in diagnosis and risk prediction. However, the rigor of these complex methods is fundamentally dependent on the quality and integrity of the underlying datasets and the validity of their statistical design. We propose an emblematic case where advanced analysis (ML/Big Data) must necessarily be subsequent to the verification of basic methodological coherence and adherence to established medical protocols, such as the STROBE Statement. This study highlights a crucial cautionary principle: sophisticated analyses amplify, rather than correct, severe methodological flaws rooted in basic design choices, leading to misleading or contradictory findings. By applying simple, standard descriptive statistical methods and established national epidemiological benchmarks to a recently published cohort study on COVID-19 vaccine outcomes and severe adverse events, like cancer, we expose multiple, statistically irreconcilable paradoxes. These paradoxes, specifically the contradictory finding of an increased cancer incidence within an exposure subgroup, concurrent with a suppressed overall Crude Incidence Rate compared to national standards, definitively invalidate the reported risk of increased cancer in the total population. We demonstrate that the observed effects are mathematical artifacts stemming from an uncorrected selection bias in the cohort construction. This analysis serves as a robust reminder that even the most complex health studies must first pass the test of basic epidemiological consistency before any conclusion drawn from subsequent advanced statistical modeling can be considered valid or publishable. We conclude that robust methods, such as Propensity Score Matching (PSM), are essential for achieving valid causal inference from administrative data in the absence of randomization, but PSM alone, even if sophisticated, is insufficient to guarantee the clinical meaning or external validity of the results if applied without regard for the cohort's epidemiological representativeness.

**Keywords:** Advanced Computational Methods in Health, Computational Epidemiology, Statistical Paradox, Selection Bias, External validity

## 1. Introduction

The integration of advanced analytical tools, including Machine Learning (ML), Big Data processing, and computational epidemiology, has fundamentally revolutionized modern health research [1, 2]. The analysis of vast repositories of electronic health records and national administrative databases promises unprecedented accuracy in diagnosis, risk prediction, and the identification of subtle health trends. These massive datasets allow for the construction of population-based cohorts large enough to detect rare events or small effects, which is crucial for evaluating widespread public health interventions.

However, the rigor of these complex methods is fundamentally dependent on the quality and integrity of the underlying datasets and the validity of their statistical design. Importantly, the accurate assessment of post-marketing serious adverse events (or SAE), particularly those associated with widespread public health interventions, like COVID-19 vaccination for example, is critical for public trust and effective health policy. Observational studies drawing from national health databases are hence essential tools in this process, offering large sample sizes and real-world data. However, the reliability of such studies is fundamentally dependent on the methodological rigor applied to cohort selection and statistical adjustment [3]. Critically, it is a cause for concern when studies emerge that fail to adhere to the basic biostatistical requirements for cohort construction, effectively rendering their results less valid or unreliable.

Within this context, a core standard of rigor is External Validity [4]. This concept refers to the extent to which the findings of a study can be generalized to other populations, settings, and circumstances outside the study's specific cohort. For example, for a cohort derived from a national registry, high external validity requires that the study's overall burden of disease (measured by the Crude Incidence Rate, or CR) is statistically consistent with the known national burden of disease for the same period. Failure to meet this standard, often due to sampling or selection issues, means the cohort is not representative of the broader population, rendering its conclusions questionable in a real-world context. Methodological integrity in observational studies is therefore expressed by means of adherence to international reporting guidelines, notably the STROBE Statement (STrengthening the Reporting of OBservational studies in Epidemiology). Specifically, Item 21 (Generalizability) requires authors to discuss the extent to which their study's results can be applied to contexts outside the specific cohort analyzed [5]. The failure to adhere to this principle, for example through a profound deviation between a cohort's CR of a given observed disease and the national average, constitutes a violation of scientific transparency. A historical case illustrating the failure of generalizability due to selection bias involves the analysis of cardiovascular risk linked to the drug Vioxx (i.e., rofecoxib), where early observational study results could not be reliably extended to the general population due to non-representative cohorts [6].

When a cohort fails to reflect the true burden of disease in the reference population, any statistical association, even if internally valid, loses its significance for public health policy. Thus, the scrutinized cohort's failure to meet basic epidemiological consistency checks is equivalent to a failure to comply with STROBE Item 21, rendering its conclusions non-generalizable.

Having stated all the above, study [7] provides a recent, retrospective, population-based cohort analysis utilizing data from some South Korean administrative database to investigate the 1-year risks of cancers associated with COVID-19 vaccination in South Korea. The study's finding, suggesting a higher rate of new cancer cases among the vaccinated population compared to the unvaccinated, is a surprising and scientifically challenging result that has yet to be fully addressed and scientifically analyzed with the required depth and urgency, especially considering the global scope of the vaccination programs [8, 9].

The present work serves as a comprehensive scrutiny that integrates two sequential computational analyses. Initially, we identified a pronounced epidemiological paradox based on raw incidence calculations derived from the study's supplementary data reported in [7]. This paradox established a significant external inconsistency between the study cohort's aggregate cancer incidence and official national statistics [10-14]. The follow-up analysis we developed has been then an integral part of the overall argument and posits a specific methodological explanation for this paradox: the likely misapplication of the Propensity Score Matching (PSM) procedure [15]. Worryingly, this procedure resulted in the final cohort severely underrepresenting the high-risk elderly demographic (>= 65 years), which constituted only 12.15% of the cohort compared to the national demographic benchmark of 18.00%. This systematic underrepresentation directly leads to a violation of STROBE Item 21 (Generalizability). The failure of the cohort to accurately reflect the true age structure and, consequently, the true baseline cancer burden of the target population has demonstrated that the study's findings are not generalizable to the broader South Korean population.

In closing, the primary objective of this paper is to quantify and demonstrate the severity of the external inconsistency observed in the scrutinized cohort of [7], thereby challenging its external validity. We then propose that the PSM procedure [15], by causing a profound structural bias, is the quantifiable methodological explanation for the numerical discrepancy between the observed cohort rate and the national benchmark. This failure highlights a crucial principle: even an algorithmically sophisticated approach, such as PSM, or refined statistical techniques that learn from data, can lead to misleading and contradictory results when the reference to methodological principles inspired by descriptive statistics and comparison with epidemiological gold standards is lost. This analysis ultimately calls into question the reliability of the study's final association results. It is essential to undertake this critical examination using the known data, as the magnitude of the finding demands the highest level of scientific scrutiny.

The remainder of this paper is structured as follows. In Section 2 we illustrate the data and methods which were used for our analysis. In Section 3 we present the results we achieved. Section 4 discusses those results, while Section 5 concludes the paper.

## 2. Materials and methods

In the following, we provide sufficient details on used data and methods to allow readers to replicate our results. This Section finally concludes with the mandatory statements required for publication, including declarations regarding Ethics of Research, Clinical Trial Registration, Funding, Conflicts of Interest, and Data Availability.

### 2.1. Data Sources and Extracted Metrics

This analysis is based entirely on publicly available, aggregated data extracted from [7] and official South Korean national health statistics as reported in [10-14]. In particular, the raw cohort figures necessary for our analysis were obtained from Table S4 ("Cumulative incidences of overall cancers in the matched cohort between vaccinated and unvaccinated individuals") from the Supplementary Material of [7]. These figures are summarized in the following Table 1.

**Table 1.** Raw Cohort Data showing the initial and final matched cohort counts, case numbers, and the Propensity Score Matching (PSM) details used in [7].

| Metric | Value |
|---|---|
| Initial Cohort Size | 8,407,849 individuals |
| Final Matched Cohort Size | 2,975,035 individuals |
| Total Cancer Cases in Matched Cohort | 12,133 cancer cases |
| Unvaccinated Group (N) | 595,007 individuals |
| Unvaccinated Group (Cases) | 1,989 cancer cases |
| Vaccinated Group (N) | 2,380,028 individuals |
| Vaccinated Group (Cases) | 10,144 cancer cases |
| Propensity Score Matching (PSM) | 1:4 Ratio |

The Official National Cancer Statistical data, including the Official Crude Incidence Rate (CR), for all cancers in South Korea were instead sourced from the Korean Central Cancer Registry for the years immediately preceding and during the study period (2020–2022) as reported in [10-13]. This data provides the robust national baseline against which the study cohort's representativeness has been tested. Furthermore, the official South Korean demographic structure, which specifies that the population aged >= 65 years constitutes 18.00% of the total in the reference year 2022, was used as the expected national baseline for assessing the cohort's age representativeness, as reported in [14].

*2.2. Definition and Calculation of Crude Incidence Rate*

The Crude Incidence Rate (CR) per 10,000 population is a fundamental epidemiological measure used here specifically to evaluate the external validity of the cohort. Unlike Age-Standardized Rates (ASRs) which adjust for age distribution to allow comparison between populations, the CR reflects the raw burden of disease in a defined population over time [15]. Most importantly, any cohort derived from a national database should possess an aggregate CR that is statistically consistent with the national average CR for the same time period. A significant deviation signals a foundational problem in the initial sampling or selection process that any given procedure used to construct the cohort would fail to correct. The CR is calculated using the established epidemiological formula:

*CR* per 10,000 = (*Number of new cases during a given period*) / (*Average population at risk during the same period*) x 10,000     (1)

This is followed by a straightforward calculation of official South Korean CR baseline [10-13], whose values for both sexes per 100,000 population were converted to a per 10,000 basis to establish the national benchmark as recorded in Table 2.

**Table 2.** Official National Crude Incidence Rates (CR) for All Cancers in South Korea per 100,000 and the derived CR per 10,000, used to establish the national average baseline for the reference period 2020–2022, sourced from [10-12].

| Year | CR per 100,000 | CR per 10,000 |
|------|----------------|---------------|
| 2020 | 482.9 | 48.29 |
| 2021 | 540.6 | 54.06 |
| 2022 | 550.2 | 55.02 |

Consequently, the official average CR baseline for all cancers for the reference period (2020–2022) can be established as the mean of these values: CR (Official Average) = 52.46 per 10,000 (Standard Deviation SD = 2.97, being assumed that the three annual CR values constitute the population of the reference period, thus the SD is calculated using N as the denominator). Finally, using the raw figures from Table S4 in the Supplementary material provided in [7], the following CRs of Table 3 are calculated using the CR equation for the cohort of interest.

**Table 3.** Calculated Crude Incidence Rates (CR) for the matched cohort of [7], showing the overall rate for the entire cohort and the rates for the segregated vaccinated and unvaccinated groups.

| Group | Calculation (New Cancer Cases / Population)/10,000 | Crude Incidence Rate (CR) |
|-------|---------------------------------------------------|---------------------------|
| CR (Cohort Overall) | (12,133 / 2,975,035) x 10,000 | 40.78 per 10,000 |
| CR (Vaccinated) | (10,144 / 2,380,028) x 10,000 | 42.63 per 10,000 |
| CR (Unvaccinated) | (1,989 / 595,007) x 10,000 | 33.43 per 10,000 |

*2.3. Hypothesis Formulation on Propensity Score Matching*

A Propensity Score Matching (1:4 PSM) procedure aims to match each individual in the Treatment Group with four comparable individuals from the Control Group [15]. In general, The Propensity Score Matching (PSM) is a (quasi-experimental) statistical method used to reduce the confounding bias that occurs when estimating the effect of a treatment or intervention (like vaccination, in our case) in observational studies. The Propensity Score is defined as the conditional probability of an individual receiving the treatment given a set of observed covariates (e.g., age, sex, comorbidities). Hence, the propensity score $e(X)$ is given by $e(X)) = Prob\ (Z = 1\ |\ X)$, where $Z$ is the treatment assignment and $X$ is the vector of baseline covariates.

The PSM calculation procedure involves a multi-step process. First, a logistic regression model is constructed to estimate the propensity score for every individual, based on the set of observed confounders. Once the propensity scores are calculated, the matching phase begins. Different matching algorithms exist (e.g., nearest neighbor, caliper, or kernel matching). In the reported 1:4 PSM, each treated individual (or the base group) should be paired with four comparable control individuals whose propensity scores are nearly identical. This process would effectively create a synthetic, balanced cohort where the two groups are comparable on all measured confounders, thereby minimizing selection bias.

The primary rationale for using PSM is to mimic the randomization process of a randomized controlled trial in non-randomized observational data. By balancing the distribution of baseline covariates between the treated and control groups, PSM aims to isolate the true effect of the treatment from the effects of confounding factors that influenced the decision to vaccinate. If the PSM is successfully implemented, any residual difference in outcome between the matched groups can be more confidently attributed to the treatment itself. A failure in the PSM process, or a misapplication like the hypothesized inversion, fundamentally undermines this rationale and reintroduces significant bias into the analysis.

However, the application of this sophisticated methodological tool in the scrutinized study [7], specifically targeting covariate balance (such as age) between the COVID-19 vaccinated and unvaccinated groups, was performed without explicit reference to the overall generalizability of the resulting cohort. By focusing only on achieving internal balance between the two exposure groups, the procedure disregarded the fundamental epidemiological requirement that the overall final cohort must retain its external validity relative to the national population. This omission constitutes a dual failure in the respect of the STROBE Statement: firstly, Item 21 (Generalizability) is violated by failing to ensure the findings are applicable beyond the study setting (evidenced by the suppressed Crude Incidence Rate); and secondly, Item 12 (Sources of Bias) is violated by failing to report the efforts made to address the resultant structural sampling bias (the severe underrepresentation of the elderly demographic) introduced by the matching procedure itself.

*2.4. Age-Stratified Data and Chi-Squared Goodness-of-Fit Test*

To formally assess the structural integrity of the final matched cohort, we utilize the age-stratified data provided in the supplementary materials (Table S4) of [7]. The full breakdown, presented in Table 4, is essential for validating the cohort's external validity against the national age structure.

**Table 4.** Age-stratified composition of the matched cohort, based on publicly available supplementary data of [7].

| Group | Total Partecipants | Total Cancer Cases | % Partecipants < 65 | % Partecipants >= 65 |
|---|---|---|---|---|
| Unvaccinated | 595,507 | 1,989 | 87.85 | 12.15 |
| Vaccinated | 2,380,028 | 10,144 | 87.85 | 12.15 |
| Total | 2,975,035 | 12,133 | 87.85 | 12.15 |

The Chi-Squared Goodness-of-Fit Test is used to formally test the null hypothesis that the age distribution of the final PSM-matched cohort is consistent with the known age distribution of the general South Korean population (82% under 65 years, 18% 65 years and older). This statistical test quantifies the magnitude of the difference between the observed frequencies in the study cohort and the expected frequencies based on the national demographic benchmark. A statistically significant result from this test indicates that the sampling or matching procedure introduced a profound structural bias, challenging the cohort's representativeness and, consequently, its external validity. The calculation uses a degree of freedom =1 (two age categories minus one degree of freedom), based on the formula:

$$Chi-squared = \sum \frac{(O-E)^2}{E} \qquad 2)$$

where $O$ is the observed frequency in the cohort and $E$ is the expected frequency based on the national demographic proportion.

2.4.1. Ethics approval of research

No humans, animals, plants have been involved in this study which uses only publicly available, aggregated data that contains no private information. Therefore, ethical approval is not required.

*2.5. Clinical trial registration*

Since this work is a critical biostatistical analysis of publicly available data from a previously published study, it does not involve any intervention, randomized allocation, or prospective collection of patient data, and therefore does not qualify as a clinical trial and was not required to be pre-registered.

*2.6. Funding*

This research received no external funding and was conducted solely by the author as part of independent scientific inquiry.

*2.7. Conflicts of Interest*

The author declares no conflicts of interest related to the content of this article.

*2.8. Data Availability*

The findings of this study are based on calculations derived exclusively from the publicly available supplementary data of the original scrutinized publication [7] and official national epidemiological statistics [10-14]. All data sources are explicitly referenced in the manuscript. Further requests can be also addressed to the author (email: marco.roccetti@unibo.it).

3. **Results**

This Section presents two types of results: first, the quantification of the Epidemiological Paradox through the comparison of the calculated Crude Incidence Rate (CR) against the national baseline; second, the numerical evidence supporting the hypothesis of Propensity Score Matching (PSM) misutilization and its impact on the age structure of the cohort achieved with the computation of the relative Chi-squared test.

*3.1. Quantification of the Epidemiological Paradox*

The comparison between the study cohort's aggregated CR and the national average CR revealed a substantial and significant downward deviation, confirming the epidemiological paradox as summarized in the following Table 5.

**Table 5.** Quantification of the Epidemiological Paradox: Comparison of the scrutinized Cohort's overall Crude Incidence Rate (CR) against the Official National Average CR, highlighting the severe downward deviation.

| Metric | Rate per 10,000 | Reference/Calculation |
|---|---|---|
| Official National Average CR (2020–2022) | 52.46 | Table 2 |
| Cohort Overall CR | 40.78 | Table 3 |
| Absolute Deviation | - 11.68 | (40.78 - 52.46) |
| Percentage Deviation | - 22.26% | (- 11.68) / 52.46 x 100 |

The paradox is summarized as follows: the study's analysis suggests an elevated cancer risk within the majority group (vaccinated CR is 27.7% higher than unvaccinated CR), which should intuitively push the overall cohort CR higher, yet the overall CR is 22.26% lower than the national baseline. This profound inconsistency represents a strong presumption of the cohort's lack of representativeness, which awaits formal refutation, though such a refutation appears mathematically challenging.

To comprehend the full impact of this deviation, one must translate these statistical discrepancies into absolute numbers, which reveal the magnitude of the effect. Based on the cohort's overall rate of 40.78 per 10,000 and applying this to South Korea's population (approx. 51.77 million inhabitants [13, 14]), the cohort rate would translate to approximately 210,873 new annual cancer cases. This is over 61,000 fewer new cases than the 271,957 derived from the official national average rate of 52.46 per 10,000 for the same population. This massive deficit in expected cases underscores the profound lack of representativeness.

Definitely the scrutinized study [7] is contradicting itself, showing concurrent increases in the vaccinated and overall decrease.

*3.2. Quantification of Structural Bias: PSM Numerical Signature and External Validity Failure*

The scrutiny of the final reported cohort sizes in [7] suggests a significant numerical anomaly in the application of the Propensity Score Matching (PSM) procedure. Given the study's focus on the COVID-19 vaccine, the standard approach would define the Vaccinated group as the Treatment group. However, the final cohort sizes reported 595,007 Unvaccinated and 2,380,028 Vaccinated, demonstrating a precise numerical correspondence: the total matched cohort (2,975,035) is exactly five times the size of the smaller unvaccinated group (595,007). This numerical construction suggests that the smaller unvaccinated pool defined the base cohort size for the matching. This numerical signature supports the hypothesis of an atypical application of the standard PSM procedure, yielding an inversed 4:1 ratio: Vaccinated / Unvaccinated. This observation provides the initial groundwork for the structural damage quantified below.

The structural consequence of the PSM procedure is fundamentally critical, as it compromised the cohort's external validity. The analysis of the age-stratified data from Table 4 already reveals a severe alteration: the cohort's total population aged >= 65 years constitutes only 12.15% of the total cohort, representing a substantial 5.85 percentage point downward deviation from the expected national demographic average of 18.00% for the >= 65 age bracket. This systematic underrepresentation of the high-risk elderly demographic is the direct consequence of the PSM procedure, which prioritized internal balance between the two exposure groups over external consistency with the national population.

To formally assess the statistical significance of this profound deviation, we performed a Chi-Squared Goodness-of-Fit Test, comparing the observed age structure of the cohort against the national benchmark (82% under 65 years vs. 18% 65 years and older). The test yielded an exceptionally high value of Chi-squared of 69,370 (with df = 1). This result translates into a *p value* significantly lower than 0.00001. This overwhelming statistical evidence confirms that the final matched cohort is not a random representation of the South Korean population and possesses a structural age distribution that is profoundly biased towards the younger, lower-risk demographic. This quantified structural defect provides the direct methodological explanation for the suppressed Crude Incidence Rate (CR) identified in the previous Section, proving that the PSM procedure, designed to reduce bias, (inadvertently) destroyed the cohort's external representativeness, as definitely summarized in Table 6 below.

**Table 6.** Summary of Structural Inconsistency.

| Metric | Scrutinized Cohort Observed Value | National Benchmark Expected Value | Inconsistency Result |
|---|---|---|---|
| Crude Incidence Rate (CR) | 40.78 per 10,000 | 52.46 per 10,000 | - 22.26% Deviation (Epidemiological Paradox) |
| Population Aged >= 65 Years | 12.15% | 18.00% | 5.85% Underrepresentation |
| Statistical Significance of Age Bias | Chi squared = 69,370 | N/A | p value < 0.00001 (Highly Significant) |

## 4. Discussion

The first part of this Discussion synthesizes the main quantitative findings, focusing on the core issues raised. First, it is worth while reminding that this research has integrated computational epidemiological analyses to critically evaluate the methodology and findings of a given scrutinized cohort from a given study [7] with a surprising finding of a higher cancer incidence in the COVID-19 vaccinated group. Our initial critique established a pronounced epidemiological paradox: while the cohort suggested a higher crude cancer incidence rate (CR) among the vaccinated group, the overall cohort CR was found to deviate downwards by over 22.26% from the official national average CR (2020–2022). This fundamental discrepancy suggested a lack of external validity (STROBE). We then found and statistically validated a plausible methodological explanation for this paradox: the reported 1:4 Propensity Score Matching (PSM) procedure resulted in a quantifiable structural bias in the cohort's age composition. This bias essentially introduces unidentified confounding factors and artificially deviate the overall CR downwards.

Essentially, our research has revealed an epidemiological paradox demonstrating a profound lack of external validity for the cohort of [7]. This is evidenced by the suppressed overall CR of 40.78 per 10,000 compared to the national average of 52.46 per 10,000. This fundamental inconsistency suggests that the sample is not representative of the underlying population's cancer incidence. Further, the discrepancy is so large that it cannot be dismissed as a minor fluctuation, calling into question the reliability of the study's conclusions. Second, the hypothesis that the 1:4 PSM was misutilized provides a concrete methodological explanation for the observed low CR. While the numerical signature initially suggested an inversion in the base group assignment, the subsequent Chi-squared test definitively quantifies the structural consequence of this procedure. The highly significant value of the Chi-squared test has proven that the PSM process, or the initial sampling it followed, resulted in a cohort that severely underrepresents the high-risk demographic group ($>= 65$ years), which makes up only 12.15% of the final sample instead of the expected 18.00%. This overwhelming statistical evidence confirms that the final matched cohort is not a random representation of the investigated population and possesses a structural age distribution that is profoundly biased towards the younger, lower-risk demographic.

This quantified alteration of the age structure, favoring a younger population with a lower baseline cancer incidence, explains the suppression of the overall CR compared to the national average. Furthermore, the systematic removal of the highest-risk demographic (aged over 65 years) undermines the entire cohort representativeness, where cancer incidence is known to be far higher (often double or more). This catastrophic failure in cohort construction prevents us from reliably assessing whether the PSM adequately balanced baseline health status and latent cancer risk even within the reduced elderly subgroup that remained in the vaccinated versus the unvaccinated cohort. Thus, the PSM procedure, intended to balance known confounders, effectively destroyed the external validity of the cohort by altering its fundamental demographic signature.

In fairness, we must also remember that, while matching from the less numerous group (unvaccinated) is standard statistical practice, in a scenario where the exposed group (vaccinated) is the overwhelming majority, in this case this choice has significantly constrained the cohort construction. Even if the PSM were executed correctly according to its internal algorithm, this choice would always drastically reduce the external validity, as the resulting cohorts would lose their ability to reflect the true epidemiological background of the original population.

It should be noticed that the present analysis study has several strengths, which we condense into three fundamental points: a) the study provides concrete, reproducible mathematical evidence (the 22.26% CR deviation) establishing a fundamental lack of external validity, which overrides subsequent statistical associations; b) it formulates a specific, testable hypothesis, the misutilized PSM procedure, which would numerically reconcile the observed epidemiological paradox, offering a concrete explanation for the bias; c) the critique shifts the focus from clinical outcomes to core methodological integrity (STROBE), serving as a crucial cautionary example for future large-scale epidemiological studies utilizing administrative data.

The limitations inherent to this present study, instead, stem primarily from the nature of observational analysis built upon aggregated, externally published data. Specifically, our quantitative findings regarding the suppressed Crude Incidence Rate (CR) and the hypothesis of Propensity Score Matching (PSM) misutilization cannot be definitively resolved without primary data access. However, these limitations are ultimately attributable to the original study, which, despite its vast scale, did not provide the necessary data access for independent validation. The ethical and scientific imperative for data transparency remains the single greatest constraint, forcing our scrutiny to rely on numerical signatures and logical inference rather than direct verification. Moreover, we must clarify that our intention is not to reject the statistical associations found in [7], showing a higher incidence of cancers in the COVID-19 vaccinated population. Such associations, while potentially valid *within the limits of their non-representative cohort*, are not the focus of this critique. However, it is impossible to ignore the profound discrepancies in the Crude Incidence Rate and the strong evidence suggesting a failure in the application of the PSM. These possible methodological flaws would render the external validity and, consequently, the reliability of the cancer risk calculation derived from this specific cohort, highly questionable until proven otherwise.

In ultimate synthesis, the case of this scrutinized cohort serves as a critical warning: the mere use of massive datasets, coupled with technically sophisticated statistical and computational techniques (such as PSM or advanced machine learning), does not automatically benefit causal inference or result in valid scientific conclusions. The value of any subsequent analysis is entirely conditional upon the integrity of the foundational methodological choices. This integrity requires unwavering adherence to basic descriptive statistics, established experimental protocols (like STROBE), comparison with epidemiological gold standards, and, ultimately, the sense of reality, that is, ensuring the study data genuinely reflects the population it aims to represent. Where these fundamental references are lost, sophisticated computational methods only amplify existing biases, leading to non-generalizable and contradictory findings.

## 5. Conclusions

The fundamental premise of any large population-based study is that the raw incidence rate of any common background event must be consistent with national epidemiological surveillance and its gold standard [16-18]. The Paradox of Crude Rates, showing simultaneously increases in the Vaccinated group and an Overall Decrease in the entire cohort, borders on nonsensical from the perspective of public health and surveillance. Specifically, the 22.26% downward deviation observed in the cohort's CR of [7], supported by our Chi-squared test quantifying the structural age bias, constitutes a plausible evidence of the lack of external validity of this exemplar case.

This failure is directly attributable to the misutilization of the Propensity Score Matching (PSM) procedure, which severely suppressed the high-risk elderly demographic in the final cohort, rendering it structurally non-representative of the national population. This methodological omission and its consequential bias represent a failure to comply with key international reporting standards: Failure of Generalizability (STROBE Item 21), as the profound structural age bias means the study's findings are fundamentally not applicable to the broader population, violating the core requirement for external validity; and Failure to Address Bias (STROBE Item 12), because the original study did not adequately report or address this systematic sampling bias introduced by the cohort construction method itself.

This failure is particularly alarming because it has driven a situation where, due to methodological and numerical inconsistencies, the signaled results cannot be accepted at a larger scale. To resolve these ambiguities and reinforce scientific rigor, the following actions are essential: a) transparency from those who should disclose the detailed algorithmic methodology used for the cohort construction, and b) full disclosure of the initial data to allow for independent verification and resolution of the raised methodological discrepancies. While public availability of the entire South Korean database is ideal, a comprehensive, extended summary of the raw data, consistent with privacy regulations, would be acceptable, provided it is detailed enough to confirm or invalidate the methodological concerns raised here. Resolving the identified methodological ambiguity is fundamental to the global trust in reported associations and broader COVID-19 vaccine safety assessments.

The scrutiny of this cohort study provides a critical precedent for the future of Big Data epidemiology. We conclude that methodological sophistication, whether through PSM, Machine Learning, or complex computational modeling, is not a substitute for fundamental epidemiological consistency. The use of sophisticated tools on non-representative data sets only serves to mask and amplify pre-existing structural biases. Our analysis, relying on simple, reproducible descriptive statistics and gold standards, exposes the critical vulnerability of such studies to a failure of external validity. Therefore, we advocate for the adoption of mandatory external validity checks as a required screening step for all large-scale observational studies using health records. This crucial step ensures that scientific rigor and the sense of reality (i.e., the cohort's representativeness) are prioritized over algorithmic complexity, safeguarding the reliability and public health utility of future research.

**Use of AI tools declaration**

The authors declare they have not used Artificial Intelligence (AI) tools in the creation of this article.

**Author Contribution**

Not applicable since there is a single author.

**Acknowledgments (All sources of funding of the study must be disclosed)**

We would like to thank our colleagues of the University of Bologna for many discussions around this subject.

**Conflict of interest**

The authors declare there is no conflict of interest.